# Siamese Network Training using Artificial Triplets by Sampling and Image Transformation

**Ammar N. Abbas**\*, **David Moser**
BSB Artificial Intelligence - OSCAR Collision Avoidance System for Boats
Linz, Austria
\* Corresponding author: ammar.abbas@eu4m.eu

*Abstract*—The device used in this work detects the objects over the surface of the water using two thermal cameras which aid the users to detect and avoid the objects in scenarios where the human eyes cannot (night, fog, etc.). To avoid the obstacle collision autonomously, it is required to track the objects in real-time and assign a specific identity to each object to determine its dynamics (trajectory, velocity, etc.) for making estimated collision predictions. In the following work, a Machine Learning (ML) approach for Computer Vision (CV) called Convolutional Neural Network (CNN) was used using TensorFlow as the high-level programming environment in Python. To validate the algorithm a test set was generated using an annotation tool that was created during the work for proper evaluation. Once validated, the algorithm was deployed on the platform and tested with the sequence generated by the test boat.

## I. Introduction

This work was carried out in a company called BSB Artificial Intelligence (BSB-AI) GmbH founded in 2017 [1]. The company focuses on technologies related to machine vision and artificial intelligence and is based in Austria, France, and Portugal. The company's first product is called Oscar – an automated collision avoidance system for ships [2]. The product focuses to address the collision risk for the boats with other ships who do not have proper navigation tools [3] or obstacles or in some cases with the marine wildlife that are not possible to be detected through radar [4], and in restricted visual scenarios where human eye fails to observe, such as through fog or at night [5]. Oscar system consists of a vision unit (including visual sensors), a user interface (called the navigation tool that displays floating objects ahead of the vessel in a defined field of view that is installed on on-board computers), and a processing unit (processes the video frames in real-time based on artificial intelligence and warns for collision risks) as shown in Figure 1. The goal was to develop a feature-based ID assignment functionality for the tracker to reduce false positives. Just as it is required by the human eye to keep track of the objects and to determine the potential dangerous obstacles that are very near or moving at high speed and towards our path, in a similar way it is required by any autonomous system to keep track of such potential dangers [6]. The main objective to keep track of the object is to determine their shape and dynamics in each frame which includes their trajectory, velocity, and acceleration that is then used to estimate their location ahead in time to avoid potential collisions. The secondary objective of the tracker is to aid the object detection algorithm when it makes a mistake in detection or misses a detection [7]. To develop such tracker three major aspects were to be considered. (i) The tracker should be online i.e. real-time because of the application requirement of warning for potential danger before the collision [8]. (ii) It should be able to track multiple objects all at once between consecutive frames without reducing too much run-time known as Multi-Object Tracking (MOT) [9]. (iii) The tracker must mimic the human vision in the sense that it can distinguish between the multiple objects based on its visible features at pixel level [10]. To solve this complex problem in a Computer Vision (CV) domain it is not possible to achieve great accuracy using traditional programming paradigms due to a high number of possible scenarios, therefore, an artificially intelligent agent had to be chosen that was trained to handle such situations [11]. The tracker was then required to be trained on our data set to learn the scenarios observed by our thermal camera. Every machine learning method must be evaluated to measure its accuracy under an unseen environment, therefore, Multi-Object Tracker Accuracy (MOTA) metric was used to evaluate the performance of the tracker on a test set that contains video frames which are different from the ones on which the algorithm is trained on to simulate the real-world scenario [12]. Once evaluated, the final task was to deploy the algorithm on the graphic card called Jetson that was responsible to do computations in real-time on the on-board system.

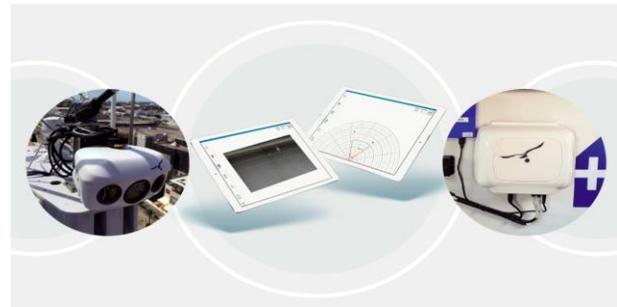

Fig. 1: Oscar system

### A. Problem statement

The vision processing unit at the current state is not able to identify and track the objects based on their features that result in a greater number of identity switches. Identity or ID switch is an error when the identity of the target object is changed after occlusion or interaction and is the core problem of object tracking that propagates the error through all the frames [13]. To minimize false positives (incorrect detections of the objects) it was necessary to elevate the performance of the tracker. At the current state, the tracking is performed using a Kalman Filter (KF) which estimates and associates IDs to individual objects. However, it fails in a high motion environment and scenarios where there are overlapping or crossing objects [14]. The current software deployed on the hardware used TensorFlow version 1.10 as the machine learning library, therefore, to integrate the tracker it had to be developed using the same library. Therefore, every algorithm adopted had to be migrated into the same format. The second challenge for the software was the limited computational capacity (memory and processing time) as computations had to be done on-board and in real-time on a single Graphical Processing Unit (GPU). Therefore, the developed algorithm must be optimized to reduce the memory required as well as it should have a high execution speed called frame rate, which is expressed in Frames per Second (FPS). FPS is a frequency at which consecutive images or frames are available for processing [15].

## II. Literature Review

Artificial Intelligence (AI) is a newly emerging field and especially its branch of Neural Networks (NN) [16]. It was important to understand the basic mathematical concepts used in machine learning and the knowledge of working with neural networks before moving towards object tracking. Further, there is a broad domain of literature that uses different neural network architectures to track the objects. However, there are several branches of object tracking under this hood that depends on the type of tracking and the type of neural network used to perform such a task. These undergo into further branching factor that includes the type of architectures developed under each domain. Therefore, a wide area of literature was studied to initialize with a basic architecture and then to modify it according to our needs. At the beginning to understand the concept of tracking and the performance of each tracker it was necessary to study some basic tracking approaches in different domains and to be able to test it. For that, a centroid-based tracker [17], Generic Object Tracking Using Regression Networks (GOTURN) [18], and Multi-Domain convolutional Neural network (MDNet) [19] were studied and tested. Based on which, basic pruning was performed based on the object tracking that was required for our system which included it to be an online, feature-based, multi-object tracker developed in TensorFlow (TF). However, it was difficult to satisfy all the conditions at once therefore MDNet was chosen to start with and later modify it according to our demands. During implementation, the MDNet was enhanced to Real-Time MDNet (RTMDNet) [20] to increase the performance of the tracker. Another approach for object tracking is known as tracking-by-detection which uses a high accuracy detector and based on the potential candidates received by detection performs tracking. Deep Simple Online and Real-Time (DeepSORT) [21] is a tracker that uses such an approach and is a potential candidate. To train DeepSORT, knowledge of the Siamese network was important that based on different image features learns to distinguish between them statistically [22]. Once a tracker was developed it is a common practice in machine learning to evaluate it based on some accuracy metrics for which the modified version of Multi-Object Tracker Accuracy (MOTA) calculation has opted [12].

## III. Methodology

This section includes the procedures used for the o object tracking.

### A. Testing platform

The current overall system called "Oscar" consists of a device that is fixed at the top of the boat's mast and functions as its vision system. It includes multiple optical sensors: two thermal cameras (for wide view range and night vision), a color camera (providing visual to the screen), an Inertial Measurement Unit (IMU) for stability and heading measurements, and a Global Positioning System (GPS) for real-time location, navigation and speed measurements. Everything connects to a central hub that is responsible for storing and manipulate sensor data and perform calculations on a graphics card used as a processing unit to display on the dashboard called the "Oscar map". The graphic card used in the hardware to deal with all the processing was the Nvidia Jetson TX2 module, which is a power-efficient AI computing device. Jetson has a 256-core Nvidia Pascal Graphical Processing Unit (GPU) and a dual-core Nvidia Denver 254-bit Control Processing Unit (CPU) with 8GB 128-bit memory and 32GB storage [23]. However, for testing algorithms, a laptop was used having an Intel Core i7, 8th generation, 1.8GHz CPU.

### B. Platform software

The current software platform is compatible with Python 3.5 and TensorFlow 1.10. Therefore, the tracking algorithm had to be developed with similar versions to avoid compatibility issues. Python is a high-level programming language and TensorFlow is an open-source library for machine learning. TensorFlow was chosen particularly because of its comprehensive, flexible ecosystem of tools, libraries, and community resources that allows building models and easy deployment on the hardware [24]. It uses a graphical-model approach where the model is built in the form of a computational graph once at the beginning and later computations are performed on that graph. This is, however, difficult for the beginners to comprehend and has a steep learning curve but is much faster, consumes lesser dynamic memory, and is easy to be deployed on various platforms. Furthermore, TensorBoard is a tool provided by TensorFlow that helps visualize the model architecture in a graphical node format and to visualize training and evaluation curves in real-time which helps in easy debugging. The processing steps connected in series, known as the pipeline, consists of acquiring the image from the thermal camera and stabilizing the image based on the calculations from IMU. After stabilization, it is passed through the neural network that detects objects in each frame of the real-time video. The network returns the class of the object along with its location on the image in the form of a bounding box. The detections are random in every frame, therefore to keep track of objects the detections are passed through Kalman Filter (KF) that estimates the position of the object in the next frame and based on the current location associates a unique ID to every object. The filter also eliminates false detections and captures missed detections. However, it fails in an environment having high motion or overlapping and crossing objects [14]. Therefore, an intelligent approach was to be adopted that could observe the physical features of the objects and reduce such problems.

### C. Object tracking

Object tracking is defined as the current state estimation of the target object from previous information [7]. To achieve it in an AI paradigm it has several categories:

- Tracking: online / offline tracking
- Principle: feature/ motion/ bounding box position
- Detection: detection-based/ detection-free
- Training: offline/ online training
- Object: single object/ multiple object
- Platform: PyTorch/ TensorFlow

To satisfy the needs, the object tracker which is required should perform online multiple-object tracking based on features of objects. An offline tracker uses the information of the previous as well as the information from the future frames which is not possible in real-time (online) tracking. It can be detection-based or detection-free. The training can be online or offline, but the time constraints and memory capacity should be kept in mind, and as mentioned earlier the platform to be used is TensorFlow. Therefore, these were the hard constraints:

- Tracking: online tracking
- Principle: feature-based
- Object: multiple objects
- Platform: TensorFlow

*1) DeepSORT:* Simple Online and Real-time Tracking with a Deep association metric (DeepSORT) [21] is an online, multi-object, and feature-based tracker developed by Wojke et al. It uses an offline Deep learning training approach which makes it a real-time tracker. DeepSORT uses the information provided by the object detection to generate track candidates and performs tracking based on the estimation provided by the Kalman Filter (KF). Moreover, in addition to the KF, it uses the information about the physical features obtained from the neural networks about the objects and assigns the identity to each track by the combination of two distance metrics called association metrics described in Equation 1, where Lambda is a weight parameter used to provide weightage to each distance (cost) matrix (Mahalanobis distance [25] providing the motion model for short term predictions and Deep Appearance Descriptor trained using Siamese triplet network [22] providing appearance information to recover identities for long term occlusions). To assign the IDs efficiently it utilized the Hungarian assignment [26] algorithm. The architecture is shown in Figure 2. DeepSORT was modified by replacing the Kalman filter with the Euclidean distance as shown in Figure 3 metrics because the KF performs poorly in a high motion/disturbance environment (non-linear). The appearance descriptor was trained using a Siamese triplet network which was combined with the Euclidean distance and the tracker provided several features:

- Initiate tracks = centroid distance threshold,
- Restore tracks = previous features,
- Forget tracks = maximum age, and
- Assignment = cost matrix threshold.

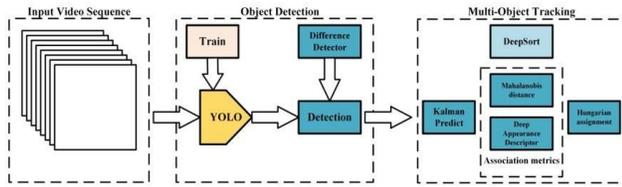

Fig. 2: DeepSORT architecture

$$D = Lambda \cdot D_k + (1 - Lambda) \cdot D_a \quad (1)$$

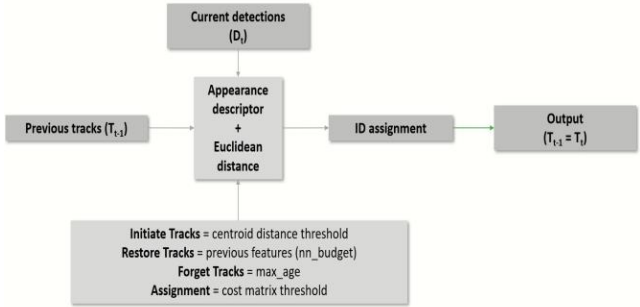

Fig. 3: Updated DeepSORT model

### D. Appearance Descriptor (Siamese Triplet Network)

The appearance descriptor is used to distinguish between the objects based on their physical features on pixel-level called embeddings using deep learning (neural network) approach. The loss function optimized to learn such features was the triplet loss. The architecture used is called the Siamese triplet network. During the learning, it uses three images the anchor (previous detection), the positive (current and future detections), and the negative (all the other objects). It is trained to minimize the distance between the anchor and the positive sample and to maximize the distance between the anchor and the negative sample as described by Equation 2, where the margin is the threshold for the separation between the positive and negative samples and for negative results it returns 0; indicating that no more effort is needed to separate those examples as the threshold has been reached. These three images are passed through a CNN layer generating a 1-dimensional embedding which is used to calculate the distances as shown in Figure 4. The optimization is performed on any one set of weights as all the weights are shared among the three layers. During inference, only one network is used which provides an embedding layer for the candidate track (anchor) and finds the best match in the current detection that has the least distance from it. The network was created in TensorFlow as shown in Figure 5. As in our case, there were no such samples available to train the network, therefore, a new approach was adopted that generates samples around the object using a certain scale and translation factor motivated by the MDNet. These samples serve to simulate as the positive future object detections of the anchor image. This approach was named "Siamese training using sampled triplets".

$$L = \max(d(a, p) - d(a, n) + margin, 0) \quad (2)$$

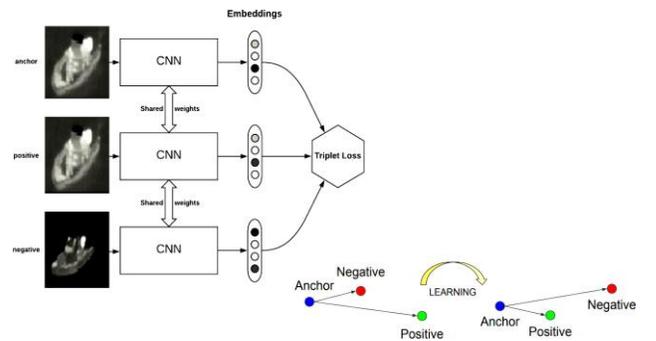

Fig. 4: Siamese triplet loss architecture

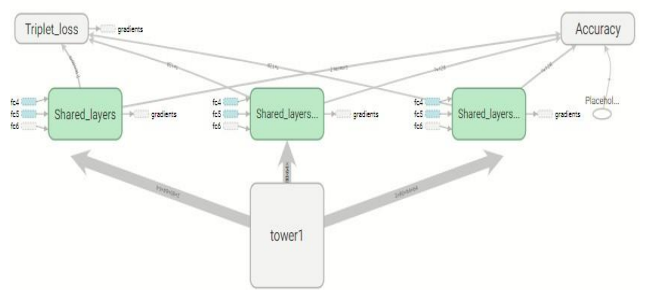

Fig. 5: Siamese triplet network TensorFlow model

*1) Architectures:* Two architectures were used and compared for the training, the first one generated the embeddings directly from the pre-trained feature extractor (object detection) network using fully connected layers, and the other used two additional convolutional layers before the final fully-connected layers to generate the embeddings with dimension 5*5 and 3*3 with channel-depth 128 and 256, respectively as shown in figure 6. The best network was chosen to perform some hyper-parameter tuning and to pick the best suitable parameters.

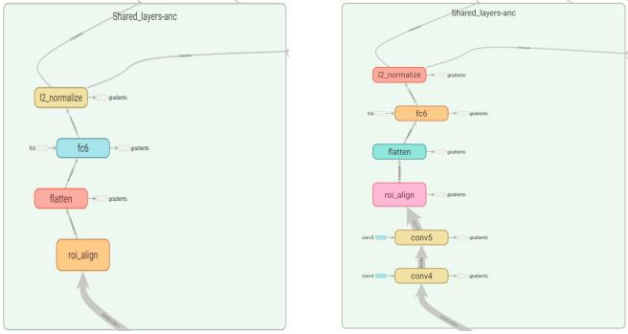

Fig. 6: Triple network training architectures; without convolution (left), with convolution (right)

*2) Image augmentation:* Some possible improvements were identified during inference. One could choose water as the negative samples to maximize the distance and remove any false detection that may occur during the real-time application. The other enhancement was to use image augmentation and perform image transformations for positive samples to simulate the future detections even better by shearing the image in either axis or rotating the object at a certain angle as the case which may be observed in the sea as shown in Figure 7. The final improvement that could be made was to annotate such a data set having future detections from the video frames and training on it, however, that may require a lot of time investment.

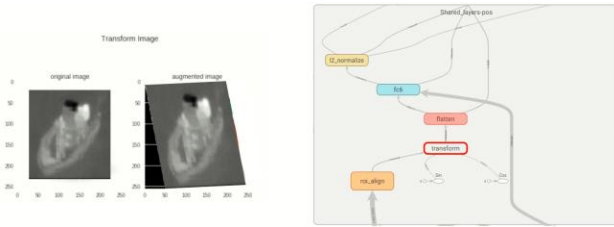

Fig. 7: Siamese Triple network with augmentation

### E. Evaluation Metrics

To evaluate the performance of the tracker it was required to have some ground-truth labels for the tracker to measure the accuracy.

*1) Annotation tool:* As no such data set was available, therefore, an annotation tool was created to aid the annotation of the tracker using a sequence. The tool displayed the previous and the current frame to help visualize the IDs in the previous frame and assign the same IDs. The initial assignment was performed using centroid distance and later with the help of the tool by clicking at any object, one could edit the ID. It was also possible to draw or delete an unwanted annotation as shown in Figure 8. Some annotations were deleted intentionally to simulate a real environment for the tracker where there are some missed detections. Some extra annotations were drawn in water to emulate the false positives as shown in Figure 9.

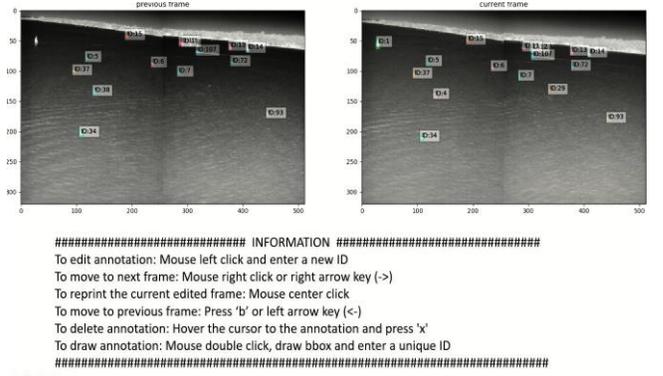

Fig. 8: Annotation tool

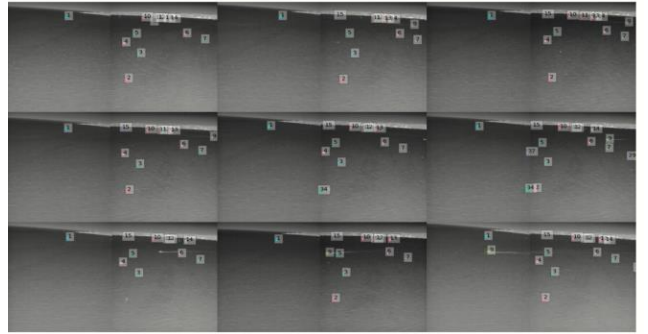

Fig. 9: Validation set

*2) Multi-Object Tracker Accuracy (MOTA) metrics::* To evaluate the performance of the tracker a modified form of accuracy metrics was utilized called Multi-Object Tracker Accuracy (MOTA) that gives a single number evaluation to compare between the tuned parameters or other trackers. MOTA calculates the average of the total number of identity switches per number of objects present in a frame for the whole sequence as described in Equation 3. The tracker was compared using different tracking parameters and for each test case, the best performing tracker was chosen and moved to the next test case.

- Case 1: based on the training model
- Case 2: based on the cost metrics
- Case 3: based on the tracker parameters

$$\text{MOTA} = \frac{\sum_{n=1}^{num\ of\ frames} 1 - \frac{num\ of\ id\ switches}{num\ of\ objects}}{num\ of\ frames} \quad (3)$$

## IV. EXPERIMENTATION

This section discusses in detail the experimentation performed based on the methodologies provided in the previous section for object tracking.

### A. Siamese Triplet network training model architectures

The way a Siamese network was trained was by using the candidate sampling strategy and using those samples as the positive examples for the anchor object and all the other samples of other objects as negative examples as shown in Figure 10. The two model architectures that were discussed in the methodology section were compared in a qualitative way by analysing the confusion matrix [27] and it was observed that the one with convolutional layers performed relatively much better as shown in Figure 11. However, the inference time was still to be examined as convolutional layers are computationally expensive.

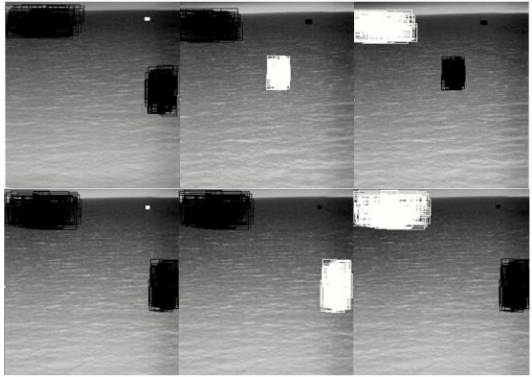

Fig. 10: Siamese Triplet network training

Fig. 11: Model architectures' comparison (just embeddings vs convolution)

*1) Inference:* Inference time for both the training architectures was compared, and not much difference was observed as shown in Figure 12, therefore, the convolutional model was selected that performed at a rate of around 0.5 FPS.

Fig. 12: Inference time comparison

*2) Hyper-parameter tuning:* After building the training model, it was tuned for performance evaluation on several hyper-parameters based on the resolution (dimension) of the ROI align layer, the depth of feature extraction layer from the detector network, alpha used as a separation margin between positive and negative samples and units of final embedding fully connected layer as shown in Table 1. All the models with different parameters were trained for 10 epochs and compared for which the following hyper-parameters performed better than the rest as shown in Figure 13:

TABLE I: Hyper-parameter tuning

| Parameters | Before | Tuning |
|---|---|---|
| Resolution (ROI) | 2 | 2 |
| Feature extraction layer | 3rd | 6th |
| Alpha (margin) | 2 | 1 |
| units of embedding layer | 128 | 512 |

Fig. 13: Hyper-parameter tuning comparison

*3) Image augmentation:* An Image transformation layer was added to the triplet model of positive samples tuned on hyper-parameters. After performing image transformation on the training data set for positive samples of the triplet network, the network performed better than the previous during inference as shown in Figure 14.

Fig. 14: Image augmentation experimentation

*4) Training logs:* After all necessary improvements to the model, the network was trained on the larger data set on a supercomputer using 11000 frames that included the video sequences recorded using our thermal camera on the boat. The training logs were visualized using TensorBoard as shown in Figure 15.

*5) MOTA evaluation:* After having the trained model weights, the tracker was tested on a validation set using MOTA evaluation metrics. As the MOTA provided a single number for evaluation, it was easier to compare various trained models and tracker parameters using the validation set and reaching a conclusion for the best tracker

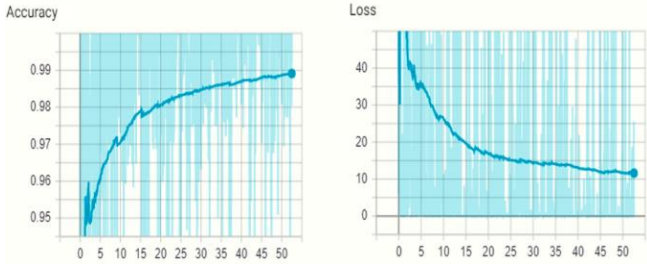

Fig. 15: Training logs

model. The evaluation was done based on the following parameters:

- Checkpoints: augmentation, shuffling, trained epochs, model parameters
- Cost matrices: appearance, distance, appearance + distance
- Tracker parameters: initialization threshold, cost matrix threshold, age of the tracker object, object features from previous frames (budget)

The tracker was evaluated on each step and the best was propagated to the next evaluation phase. The final tracker accuracy performance was 93% as shown in Figure 16, 17 and 18.

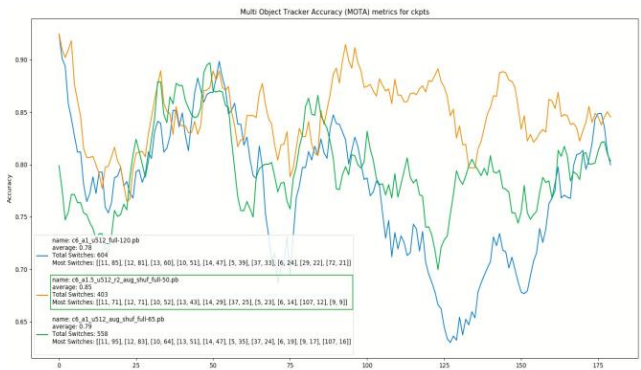

Fig. 16: MOTA for checkpoints

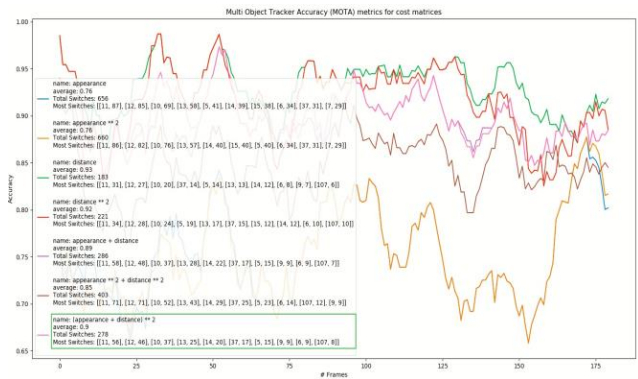

Fig. 17: MOTA for cost matrices

## V. RESULTS

The tracker was integrated into the product's platform software and was tested for its integrity. The tracker required input from the detector network for that reason, two models were created one performing detection and giving input to the tracker and other for the tracker network. The model was converted into a deployable file with frozen weights to minimize size, increasing efficiency, and easy deployment and production-ready [28]. The tracker was tested on a sequence received by the thermal camera on the boat as shown in Figure 20 and run-time was measured to be 29ms, which was fast enough for real-time tracking application. However, there were some ID switches at the section where two images from thermal cameras were fused. The tracker was also unable to restore the ID of the objects that were reentered in the frame. In some cases, the tracker lost confidence for the object and did not provide any ID to the object through some frames. The possible solution is to retrain the network on a larger data set with regularization to avoid over-fitting (generalize) and learn about the domain features more robustly.

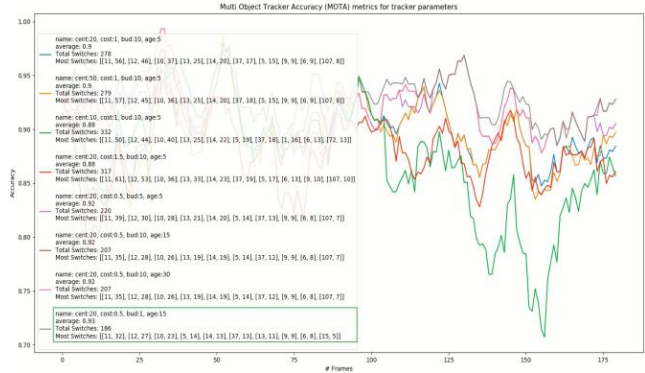

Fig. 18: MOTA for tracker parameters

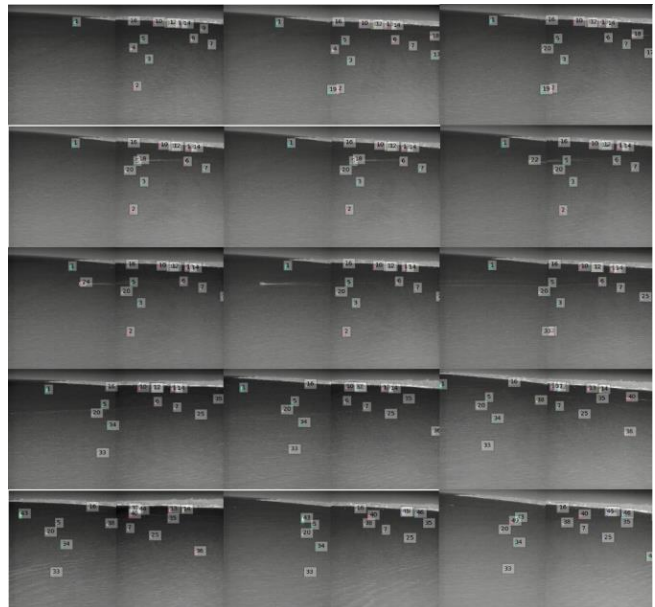

Fig. 19: Multiple objects tracking validation

## VI. CONCLUSION AND FUTURE WORK

Siamese training without using actual triplets could be improved by using adaptive ROI align strategy as shown in Figure 21(a) [29], where the interpolation is performed according to the size of the object hence preserving as much information as it can, and by training using Cosine distance instead of Euclidean distance between the triplets as shown in Figure 21(b) [30]. The cosine distance gives the separation angle between the object's embedding that makes it independent of the magnitude of embeddings.

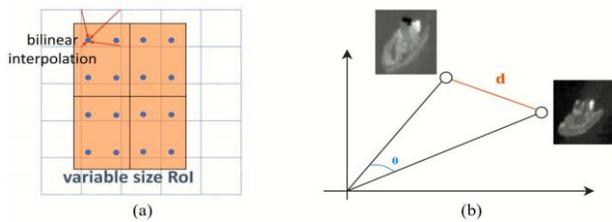

Fig. 20: (a) Adaptive ROI align, (b) Cosine distance

### ACKNOWLEDGMENTS

This work was conducted in a company named BSB Artificial Intelligence GmbH, Austria and under the supervision of two academic institutions namely École Nationale Supérieure de Mécanique et des Microtechniques (ENSMM) de Besancon, France, and Escuela Politécnica de Ingeniería (EPI) de Gijón, Spain. Therefore, I would like to thank all the institutions involved in providing the means to complete the thesis work that achieved high-quality results. Specifically.